\documentclass[11pt,a4paper]{article}
\usepackage[hyperref]{acl2018}
\usepackage{times}
\usepackage{latexsym}

\usepackage{times}
\usepackage{latexsym}
\usepackage{tabularx}
\usepackage{microtype}
\usepackage{graphicx}
\usepackage{paralist}
\usepackage{subfig}
\usepackage[utf8]{inputenc}
\usepackage{amsmath}
\usepackage{amssymb}
\makeatletter
\newcommand\footnoteref[1]{\protected@xdef\@thefnmark{\ref{#1}}\@footnotemark}
\makeatother
\usepackage{mathtools}
\usepackage{multirow}
\usepackage{booktabs}
\usepackage{xspace}
\aclfinalcopy 
\def\aclpaperid{421} 

\newcommand{\R}{\mathbb{R}}
\DeclareMathOperator*{\argmin}{arg\,min}
\newcommand{\Csource}{C_{\textrm{source}}}
\newcommand{\Ctarget}{C_{\textrm{target}}}

\newcommand{\mt}{\textsc{Mt}\xspace}
\newcommand{\blse}{\textsc{Blse}\xspace}
\newcommand{\barista}{\textsc{Barista}\xspace}
\newcommand{\artetxe}{\textsc{Artetxe}\xspace}
\newcommand{\mono}{\textsc{Mono}\xspace}
\newcommand{\ensemble}{\textsc{Ensemble}\xspace}

\newcommand{\rt}[1]{\rotatebox{90}{#1}}

\newcommand{\eg}{\textit{e.\,g.}\xspace}

\newcommand{\F}{$\text{F}_1$\xspace}

\title{Bilingual Sentiment Embeddings: \\ Joint Projection of Sentiment Across Languages}

\author{%
  Jeremy Barnes,
  Roman Klinger, \and
  Sabine Schulte im Walde\\
  Institut f\"ur Maschinelle Sprachverarbeitung\\ University of
  Stuttgart\\ Pfaffenwaldring 5b, 70569 Stuttgart, Germany\\
  \texttt{\{barnesjy,klinger,schulte\}@ims.uni-stuttgart.de}
}

\date{}

\begin{document}
\maketitle
\begin{abstract}
  Sentiment analysis in low-resource languages suffers from a lack of
  annotated corpora to estimate high-performing models.  Machine
  translation and bilingual word embeddings provide some relief
  through cross-lingual sentiment approaches. However, they either
  require large amounts of parallel data or do not sufficiently
  capture sentiment information. We introduce Bilingual Sentiment
  Embeddings (\blse), which jointly represent sentiment information in
  a source and target language.  This model only requires a small
  bilingual lexicon, a source-language corpus annotated for sentiment,
  and monolingual word embeddings for each language. We perform
  experiments on three language combinations (Spanish, Catalan,
  Basque) for sentence-level cross-lingual sentiment classification
  and find that our model significantly outperforms state-of-the-art
  methods on four out of six experimental setups, as well as capturing
  complementary information to machine translation. Our analysis of
  the resulting embedding space provides evidence that it represents
  sentiment information in the resource-poor target language without
  any annotated data in that language.
\end{abstract}

\section{Introduction}
\label{intro}
Cross-lingual approaches to sentiment analysis are motivated by the
lack of training data in the vast majority of languages. Even
languages spoken by several million people, such as Catalan, often
have few resources available to perform sentiment analysis in specific
domains. We therefore aim to harness the knowledge previously collected
in resource-rich languages.

Previous approaches for cross-lingual sentiment analysis typically
exploit machine translation based methods or multilingual
models. Machine translation (\mt) can provide a way to transfer
sentiment information from a resource-rich to resource-poor languages
\cite{Mihalcea2007,Balahur2014d}. However,  \mt-based methods require large
parallel corpora to train the translation system, which are often not
available for under-resourced languages.

Examples of multilingual methods that have been applied to
cross-lingual sentiment analysis include domain adaptation methods
\cite{Prettenhofer2011b}, delexicalization \cite{Almeida2015}, and
bilingual word embeddings
\cite{Mikolov2013translation,Hermann2014,Artetxe2016}. These
approaches however do not incorporate enough sentiment information to
perform well cross-lingually, as we will show later.

We propose a novel approach to incorporate sentiment information in a
model, which does not have these disadvantages. \emph{Bilingual
  Sentiment Embeddings} (\blse) are embeddings that are jointly
optimized to represent both (a) semantic information in the source and
target languages, which are bound to each other through a small
bilingual dictionary, and (b) sentiment information, which is
annotated on the source language only. We only need three resources:
(i) a comparably small bilingual lexicon, (ii) an annotated sentiment
corpus in the resource-rich language, and (iii) monolingual word
embeddings for the two involved languages.

We show that our model outperforms previous state-of-the-art models in
nearly all experimental settings across six benchmarks. In addition,
we offer an in-depth analysis and demonstrate that our model is aware
of sentiment. Finally, we provide a qualitative analysis of the joint
bilingual sentiment space. Our implementation is publicly available at
\url{https://github.com/jbarnesspain/blse}.

\section{Related Work}
\label{relatedwork}

\textbf{Machine Translation:} Early work in cross-lingual sentiment
analysis found that machine translation (\mt) had reached a point of
maturity that enabled the transfer of sentiment across languages.
Researchers translated sentiment lexicons \cite{Mihalcea2007,Meng2012}
or annotated corpora and used word alignments to project sentiment
annotation and create target-language annotated corpora
\cite{Banea2008,Duh2011a,Demirtas2013,Balahur2014d}.

Several approaches included a multi-view representation of the data
\cite{Banea2010,Xiao2012} or co-training \cite{Wan2009,Demirtas2013}
to improve over a naive implementation of machine
translation, where only the translated data is used. There are also approaches
which only require parallel data \cite{Meng2012,Zhou2016,Rasooli2017},
instead of machine translation.

All of these approaches, however, require large amounts of parallel
data or an existing high quality translation tool, which are not
always available.
A notable exception is the approach proposed by \newcite{Chen2016}, an
adversarial deep averaging network, which trains a joint feature
extractor for two languages. They minimize the difference between
these features across languages by learning to fool a language
discriminator, which requires no parallel data. It does, however,
require large amounts of unlabeled data.

\paragraph{Bilingual Embedding Methods:} Recently proposed bilingual
embedding methods \cite{Hermann2014,Chandar2014,Gouws2015} offer a
natural way to bridge the language gap. These particular approaches to
bilingual embeddings, however, require large parallel corpora in order
to build the bilingual space, which are not available for all language
combinations.

An approach to create bilingual embeddings that has a less prohibitive
data requirement is to create monolingual vector spaces and then learn
a projection from one to the other. \newcite{Mikolov2013translation}
find that vector spaces in different languages have similar
arrangements. Therefore, they propose a linear projection which
consists of learning a rotation and scaling
matrix. \newcite{Artetxe2016, Artetxe2017} improve upon this approach
by requiring the projection to be orthogonal, thereby preserving the
monolingual quality of the original word vectors.

Given source embeddings $S$, target embeddings~$T$, and a bilingual
lexicon $L$, \newcite{Artetxe2016} learn a projection matrix $W$ by
minimizing the square of Euclidean distances
\begin{equation}
\argmin_W \sum_{i} ||S'W-T'||_{F}^{2}\,,
\end{equation}
where $S' \in S$ and $T' \in T$ are the word embedding matrices for
the tokens in the bilingual lexicon $L$. This is solved using the
Moore-Penrose pseudoinverse $S'^{+} = (S'^{T}S')^{-1}S'^{T}$ as $ W =
S'^{+}T'$, which can be computed using SVD. We refer to this approach
as \artetxe.

\newcite{Gouws2015taskspecific} propose a method to create a
pseudo-bilingual corpus with a small task-specific bilingual lexicon,
which can then be used to train bilingual embeddings (\barista). This
approach requires a monolingual corpus in both the source and target
languages and a set of translation pairs. The source and target
corpora are concatenated and then every word is randomly kept or
replaced by its translation with a probability of 0.5. Any kind of
word embedding algorithm can be trained with this pseudo-bilingual
corpus to create bilingual word embeddings.

These last techniques have the advantage of requiring relatively
little parallel training data while taking advantage of larger amounts
of monolingual data. However, they are not optimized for sentiment.
\\[12pt]
\noindent
\textbf{Sentiment Embeddings:} \newcite{Maas2011} first explored the
idea of incorporating sentiment information into semantic word
vectors. They proposed a topic modeling approach similar to latent
Dirichlet allocation in order to collect the semantic information in
their word vectors. To incorporate the sentiment information, they
included a second objective whereby they maximize the probability of
the sentiment label for each word in a labeled document.

\newcite{Tang2014} exploit distantly annotated tweets to
create Twitter sentiment embeddings. To incorporate distributional
information about tokens, they use a hinge loss and maximize the
likelihood of a true $n$-gram over a corrupted $n$-gram. They include a
second objective where they classify the polarity of the tweet given
the true $n$-gram. While these techniques have proven useful, they are
not easily transferred to a cross-lingual setting.

\newcite{Zhou2015} create bilingual sentiment embeddings by
translating all source data to the target language and vice versa. This 
requires the existence of a machine translation system, which is a
prohibitive assumption for many under-resourced languages, especially
if it must be open and freely accessible. This motivates approaches which can
use smaller amounts of parallel data to achieve similar results.

\begin{figure*}
\centering
\includegraphics[width=0.9\textwidth]{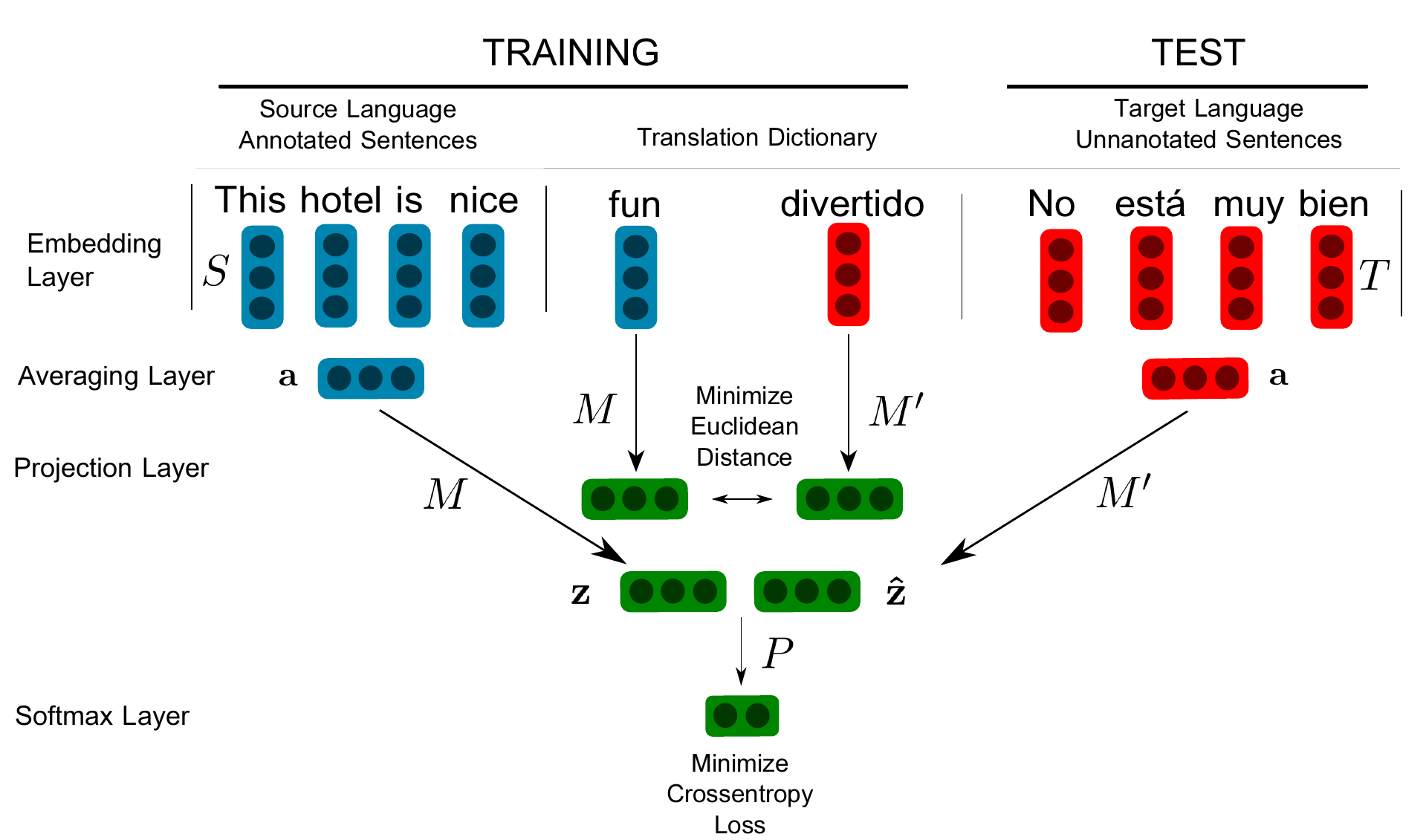}
\caption{Bilingual Sentiment Embedding Model (\blse)}
\label{fig:model}
\end{figure*}

\section{Model}

In order to project not only semantic similarity and relatedness but
also sentiment information to our target language, we propose a new
model, namely \emph{Bilingual Sentiment Embeddings} (BLSE), which
jointly learns to predict sentiment and to minimize the distance
between translation pairs in vector space. We detail the projection
objective in Section~\ref{crosslingual}, the sentiment objective in
Section~\ref{sentiment}, and the full objective in
Section~\ref{joint}. A sketch of the model is depicted in
Figure~\ref{fig:model}.

\subsection{Cross-lingual Projection}
\label{crosslingual}

We assume that we have two precomputed vector spaces $S = \R^{v \times
  d}$ and $T = \R^{v' \times d'}$ for our source and target languages,
where $v$ ($v'$) is the length of the source vocabulary (target
vocabulary) and $d$ ($d'$) is the dimensionality of the embeddings.
We also assume that we have a bilingual lexicon $L$ of length $n$
which consists of word-to-word translation pairs $L$ = $\{(s_{1},t_{1}),
(s_{2},t_{2}),\ldots, (s_{n}, t_{n})\}$ which map from source to
target.

In order to create a mapping from both original vector spaces $S$ and
$T$ to shared sentiment-informed bilingual spaces $\mathbf{z}$ and
$\mathbf{\hat{z}}$, we employ two linear projection matrices, $M$ and
$M'$. During training, for each translation pair in $L$, we first look
up their associated vectors, project them through their associated
projection matrix and finally minimize the mean squared error of the
two projected vectors. This is very similar to the approach taken by
\newcite{Mikolov2013translation}, but includes an additional target
projection matrix.

The intuition for including this second matrix is that a single
projection matrix does not support the transfer of sentiment
information from the source language to the target language. Without
$M'$, any signal
coming from the sentiment classifier (see Section \ref{sentiment}) would
have no affect on the target embedding space $T$, and optimizing
$M$ to predict sentiment and projection would only be
detrimental to classification of the target language. We analyze this further
in Section \ref{analysism}. Note that
in this configuration, we do not need to update the original vector
spaces, which would be problematic with such small training data.

The projection quality is ensured by minimizing the mean squared
error\footnote{We omit parameters in equations for better
  readability.}\footnote{We also experimented with cosine distance,
  but found that it performed worse than Euclidean distance.}
\begin{equation}
\textrm{MSE} = \dfrac{1}{n} \sum_{i=1}^{n} (\mathbf{z_{i}} - \mathbf{\hat{z}_{i}})^{2}\,,
\end{equation}
where $\mathbf{z_{i}} = S_{s_{i}} \cdot M$ is the dot product of the embedding for source word $s_{i}$ and the source projection matrix and $\mathbf{\hat{z}_{i}} = T_{t_{i}} \cdot M'$ is the same for the target word $t_{i}$.

\subsection{Sentiment Classification}
\label{sentiment}

We add a second training objective to optimize the projected source
vectors to predict the sentiment of source phrases. This inevitably
changes the projection characteristics of the matrix $M$, and
consequently $M'$ and encourages $M'$ to learn to predict sentiment
without any training examples in the target language.

To train $M$ to predict sentiment, we require a source-language
corpus $\Csource = \{(x_{1}, y_{1}),
(x_{2}, y_{2}), \ldots, (x_{i}, y_{i})\}$ where each sentence $x_{i}$ 
is associated with a label $y_{i}$.

For classification, we use a two-layer feed-forward averaging network,
loosely following \newcite{Iyyer2015}\footnote{Our model employs a
  linear transformation after the averaging layer instead of including
  a non-linearity function. We choose this architecture because the
  weights $M$ and $M'$ are also used to learn a linear cross-lingual
  projection.}. For a sentence $x_{i}$ we take the word embeddings
from the source embedding $S$ and average them to $\mathbf{a}_{i} \in
\R^{d}$. We then project this vector to the joint bilingual space
$\mathbf{z}_{i} = \mathbf{a}_{i} \cdot M$. Finally, we pass
$\mathbf{z}_{i}$ through a softmax layer $P$ to get our prediction
$\hat{y}_{i} = \textrm{softmax} ( \mathbf{z}_{i} \cdot P)$.

To train our model to predict sentiment, we minimize the cross-entropy
error of our predictions\\[-\baselineskip]
\begin{equation} 
H = - \sum_{i=1}^{n} y_{i} \log \hat{y_{i}} - (1 - y_{i}) \log (1 - \hat{y_{i}})\,.
\end{equation}

\subsection{Joint Learning}
\label{joint}
In order to jointly train both the projection component and the
sentiment component, we combine the two loss functions to optimize the
parameter matrices $M$, $M'$, and $P$ by
\begin{equation}
J =\kern-5mm
\sum_{(x,y) \in \Csource}\kern-1mm \sum_{(s,t) \in L}\kern-1mm \alpha H(x,y)
+ (1 - \alpha) \cdot \textrm{MSE}(s,t)\,,
\end{equation}
where $\alpha$ is a hyperparameter that weights sentiment loss vs.\ projection loss.

\begin{table}[tb]
\centering
\begin{tabular}{lrrrrr}
\toprule
    & & \multicolumn{1}{c}{EN} & \multicolumn{1}{c}{ES} & \multicolumn{1}{c}{CA} & \multicolumn{1}{c}{EU}\\
\cmidrule(rl){2-2}\cmidrule(l){3-3}\cmidrule(l){4-4}\cmidrule(l){5-5}\cmidrule(l){6-6}
 \multirow{3}{*}{\rt{Binary}}
 &$+$   & 1258 & 1216 & 718  & 956    \\
 &$-$   & 473 & 256 & 467  & 173   \\
 &\textit{Total}    &1731 & 1472  &   1185   & 1129        \\
\cmidrule(rl){2-2}\cmidrule(l){3-3}\cmidrule(l){4-4}\cmidrule(l){5-5}\cmidrule(l){6-6}
 \multirow{5}{*}{\rt{4-class}}
 &$++$   & 379 & 370  & 256  & 384 \\
 &$+$    & 879 & 846  & 462   & 572 \\
 &$-$    & 399 & 218  & 409   & 153 \\
 &$--$   &  74 & 38   & 58    & 20  \\
 &\textit{Total}     & 1731  & 1472     & 1185      &   1129  \\
\bottomrule
\end{tabular}
\caption{Statistics for the OpeNER English (EN) and Spanish (ES) 
as well as the MultiBooked Catalan (CA) and Basque (EU) datasets.}
\label{datasetstats}
\end{table}

\subsection{Target-language Classification}
For inference, we classify sentences from a target-language corpus
$\Ctarget$.
As in the training procedure, for each sentence, we take the word
embeddings from the target embeddings $T$ and average them to
$\mathbf{a}_{i} \in \R^{d}$. We then project this vector to the joint
bilingual space $\mathbf{\hat{z}}_{i} = \mathbf{a}_{i} \cdot
M'$. Finally, we pass $\mathbf{\hat{z}}_{i}$ through a softmax layer
$P$ to get our prediction $\hat{y}_{i} = \textrm{softmax} (
\mathbf{\hat{z}}_{i} \cdot P)$.

\section{Datasets and Resources}

\subsection{OpeNER and MultiBooked}
To evaluate our proposed model, we conduct experiments using four
benchmark datasets and three bilingual combinations. We use the OpeNER
English and Spanish datasets \cite{Agerri2013} and the MultiBooked
Catalan and Basque datasets \cite{Barnes2018}. All
datasets contain hotel reviews which are annotated for aspect-level
sentiment analysis. The labels include \textit{Strong Negative}
($--$), \textit{Negative} ($-$), \textit{Positive} ($+$), and
\textit{Strong Positive} ($++$). We map the aspect-level annotations
to sentence level by taking the most common label and remove instances
of mixed polarity. We also create a binary setup by combining the
strong and weak classes. This gives us a total of six experiments. The
details of the sentence-level datasets are summarized in
Table~\ref{datasetstats}.
\begin{table}[tb]
\centering
\begin{tabular}{lrrr}
\toprule
   & \multicolumn{1}{c}{Spanish} & \multicolumn{1}{c}{Catalan} & \multicolumn{1}{c}{Basque}\\
\cmidrule(r){1-1}\cmidrule(rl){2-2}\cmidrule(l){3-3}\cmidrule(l){4-4}
 Sentences & 23 M  &    9.6 M   &   0.7 M  \\
 Tokens  & 610 M & 183 M  & 25 M \\
 Embeddings & 0.83 M &  0.4 M & 0.14 M \\
\bottomrule
\end{tabular}
\caption{Statistics for the Wikipedia corpora and monolingual vector spaces.}
\label{stats:wikis}
\end{table}
For each of the experiments, we take 70 percent of the data for training, 20 percent for testing and the remaining 10 percent are used as development data for tuning.

\subsection{Monolingual Word Embeddings}
\label{embeddings}
For \blse, \artetxe, and \mt, we require monolingual vector
spaces for each of our languages. For English, we use the publicly
available GoogleNews
vectors\footnote{\label{note1}\url{https://code.google.com/archive/p/word2vec/}}. For
Spanish, Catalan, and Basque, we train skip-gram embeddings using the
Word2Vec toolkit\footnoteref{note1} with 300 dimensions, subsampling
of $10^{-4}$, window of 5, negative sampling of~15 based on a 2016
Wikipedia
corpus\footnote{\url{http://attardi.github.io/wikiextractor/}}
(sentence-split, tokenized with IXA pipes \cite{Agerri2014} and
lowercased). The statistics of the Wikipedia corpora are given in
Table \ref{stats:wikis}.

\begin{figure}
\centering
\includegraphics[width = \linewidth]{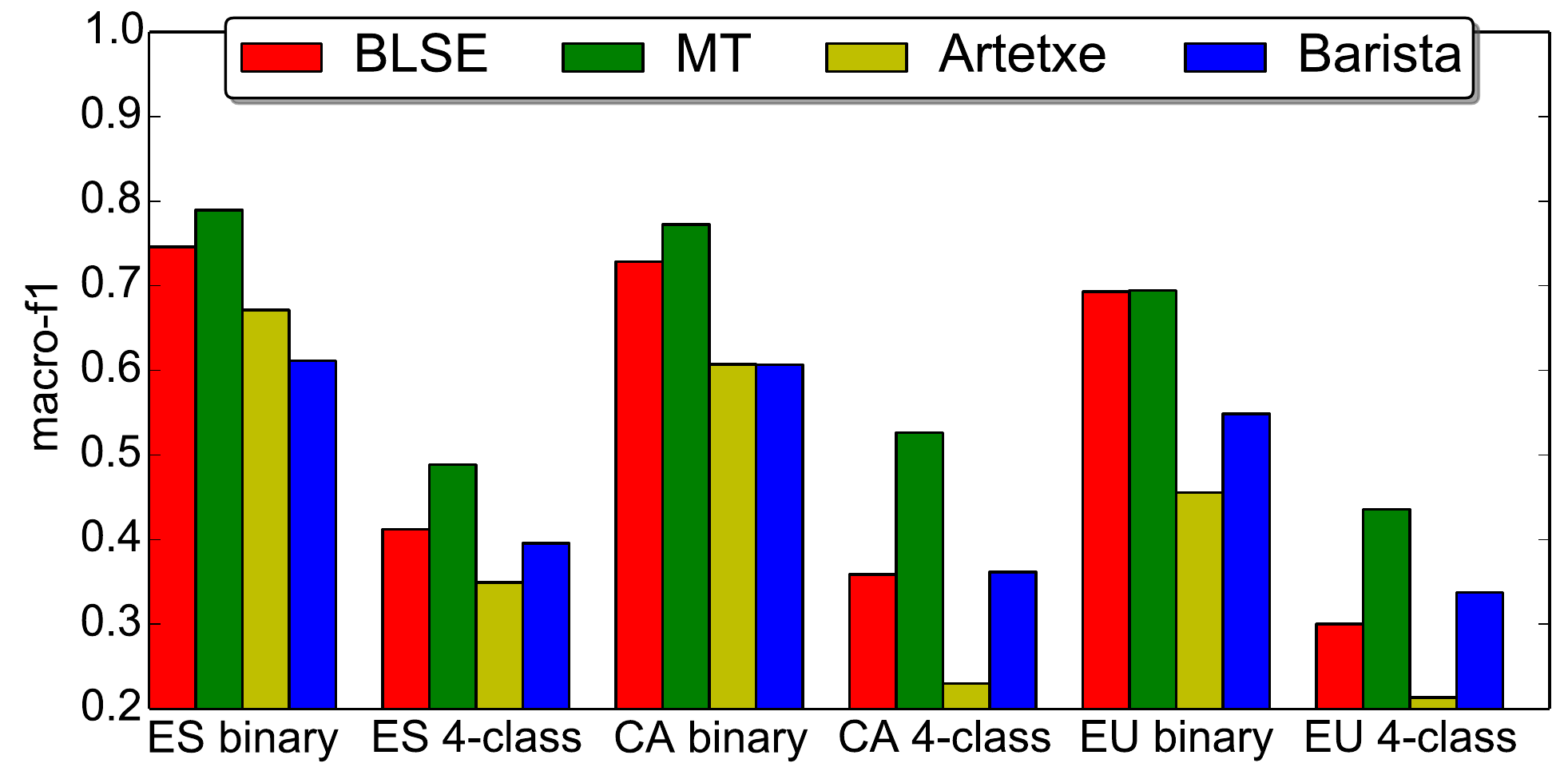}
\caption{Binary and four class macro \F on Spanish (ES), Catalan (CA), and Basque (EU).}
\label{fig:results}
\end{figure}

\subsection{Bilingual Lexicon}
\label{transdict}
For \blse, \artetxe, and \barista, we also require a bilingual
lexicon. We use the sentiment lexicon from \newcite{Huandliu2004} (to
which we refer in the following as Bing Liu) and its translation into
each target language. We translate the lexicon using Google Translate
and exclude multi-word expressions.\footnote{Note that we only do that
  for convenience. Using a machine translation service to generate
  this list could easily be replaced by a manual translation, as the
  lexicon is comparably small.} This leaves a dictionary of 5700
translations in Spanish, 5271 in Catalan, and 4577 in Basque. We set
aside ten percent of the translation pairs as a development set in
order to check that the distances between translation pairs not seen
during training are also minimized during training.

\section{Experiments}
\subsection{Setting}
\label{setting}
We compare \blse (Sections \ref{crosslingual}--\ref{joint}) to
\artetxe (Section~\ref{relatedwork}) and \barista
(Section~\ref{relatedwork}) as baselines, which have similar data
requirements and to machine translation (\mt) and monolingual (\mono)
upper bounds which request more resources.  For all models (\mono,
\mt, \artetxe, \barista), we take the average of the word embeddings
in the source-language training examples and train a linear
SVM\footnote{LinearSVC implementation from scikit-learn.}. We report
this instead of using the same feed-forward network as in \blse as it
is the stronger upper bound. We choose the parameter $c$ on the target
language development set and evaluate on the target language test set.

\begin{table}[b]
\definecolor{blue}{cmyk}{0.2,0,0,0.1}
\newcommand{\hl}[1]{{\textcolor{violet}{\emph{#1}}}}
\newcommand{\best}[1]{\textbf{\setlength{\fboxsep}{1pt}\fbox{#1}}}
\newcommand{\sep}{\cmidrule(r){3-3}\cmidrule(lr){4-6}\cmidrule(l){7-9}}
\newcommand{\sepd}{\cmidrule(lr){4-6}\cmidrule(l){7-9}}
 \centering
 \renewcommand*{\arraystretch}{1.0}
\setlength\tabcolsep{0.5mm}
 \begin{tabular}{cclrrrrrr}
 \toprule
 & & & \multicolumn{3}{c}{Binary} & \multicolumn{3}{c}{4-class}\\
 \sepd
 & & & ES & CA & EU & ES & CA & EU\\
\cmidrule(lr){4-6}\cmidrule(lr){7-9}
\multirow{7}{*}{{\rt{Upper Bounds}}} &\multirow{3}{*}{{\small\rt{\mono}}}
& P  & 75.0 & 79.0 & 74.0 & 55.2 & 50.0 & 48.3 \\
& & R  & 72.3 & 79.6 & 67.4 & 42.8 & 50.9 & 46.5 \\
& & \F & 73.5 & 79.2 & 69.8 & 45.5 & 49.9 & 47.1 \\[-0.5ex]
\cmidrule(r){3-3}\cmidrule(lr){4-6}\cmidrule(l){7-9}
 & \multirow{3}{*}{{\small\rt{\mt}}}
& P & \textbf{82.3} & 78.0 & 75.6 & 51.8 & \textbf{58.9} &43.6\\
 & & R & 76.6 & 76.8 & 66.5 & 48.5 & 50.5 & 45.2\\
 & & \F & 79.0 & 77.2 & 69.4 & 48.8 & 52.7 & 43.6\\[-0.5ex]
 \cmidrule(r){0-2}\cmidrule(lr){4-6}\cmidrule(l){7-9}

& \multirow{3}{*}{{\small\rt{\textbf{BLSE}}}} & P & 72.1 & \hl{**72.8} & \hl{**67.5} & \textbf{\hl{**60.0}} & \hl{38.1} & \hl{*42.5} \\
& & R & \textbf{\hl{**80.1}} & \hl{**73.0} & \textbf{\hl{**72.7}} & \hl{*43.4} & 38.1 & \hl{37.4} \\
& & \F & \hl{**74.6} & \hl{**72.9} & \hl{**69.3} & \hl{*41.2} & 35.9 & 30.0 \\ 
 \cmidrule(r){0-2}\cmidrule(lr){4-6}\cmidrule(l){7-9}

\multirow{6}{*}{{\rt{Baselines}}}&
\multirow{3}{*}{{\small\rt{Artetxe}}}
& P & 75.0 & 60.1 & 42.2 & 40.1 & 21.6 & 30.0\\
& & R & 64.3 & 61.2 & 49.5 & 36.9 & 29.8 & 35.7\\
& & \F & 67.1 & 60.7 & 45.6 & 34.9 & 23.0 & 21.3\\
 \sep 
&\multirow{3}{*}{{\small\rt{Barista}}}
& P & 64.7 & 65.3 & 55.5 & 44.1 & 36.4 & 34.1 \\
& & R & 59.8 & 61.2 & 54.5 & 37.9 & 38.5 & 34.3\\
& & \F & 61.2 & 60.1 & 54.8 & 39.5 & 36.2 & 33.8\\
 \cmidrule(r){0-2}\cmidrule(lr){4-6}\cmidrule(l){7-9}
 \multirow{10}{*}{{\rt{Ensemble}}}
&\multirow{3}{*}{{\small\rt{Artetxe}}} & P & 65.3 & 63.1 & 70.4 & 43.5 & 46.5 & 50.1 \\
& & R & 61.3 & 63.3 & 64.3 & 44.1 & 48.7 & 50.7 \\
& & \F & 62.6 & 63.2 & 66.4 & 43.8 & 47.6 & 49.9 \\
\cmidrule(lr){4-6}\cmidrule(l){7-9}
&\multirow{3}{*}{{\small\rt{Barista}}} & P & 60.1 & 63.4 & 50.7 & 48.3 & 52.8 & 50.8\\
& & R & 55.5 & 62.3 & 50.4 & 46.6 & 53.7 & 49.8 \\
& & \F & 56.0 & 62.5 & 49.8 & 47.1 & 53.0 & 47.8\\
\cmidrule(lr){4-6}\cmidrule(l){7-9}
&\multirow{3}{*}{{\small\rt{BLSE}}} & P & 79.5 & \textbf{84.7} & \textbf{80.9} & 49.5 & 54.1 & \textbf{50.3} \\
& & R & \textbf{78.7} & \textbf{85.5} & 69.9 & \textbf{51.2} & \textbf{53.9} & \textbf{51.4}\\
& & \F & \textbf{80.3} & \textbf{85.0} & \textbf{73.5} & \textbf{50.3} & \textbf{53.9} & \textbf{50.5}
\\
 \bottomrule
 \end{tabular}
 \caption{Precision (P), Recall (R), and macro \F of four models
 trained on English and tested on Spanish (ES), Catalan (CA), and
 Basque (EU). The \textbf{bold} numbers show the best results for
 each metric per column and the \hl{highlighted} numbers show where
 \blse is better than the other projection methods, \artetxe and
 \barista (** p $<$ 0.01, * p $<$ 0.05).}
 \label{results}
\end{table}

\textbf{Upper Bound \mono.} We set an empirical upper bound by
training and testing a linear SVM on the target language data. As
mentioned in Section~\ref{setting}, we train the model on the averaged
embeddings from target language training data, tuning the $c$
parameter on the development data. We test on the target language test
data.

\textbf{Upper Bound \mt.}
To test the effectiveness of machine translation, we translate all of
the sentiment corpora from the target language to English using the
Google Translate
API\footnote{\url{https://translate.google.com}}. Note that this
approach is not considered a baseline, as we assume not to have
access to high-quality machine translation for low-resource languages
of interest.

\textbf{Baseline \artetxe.}
We compare with the approach proposed by \newcite{Artetxe2016} which
has shown promise on other tasks, such as word similarity. In order to
learn the projection matrix $W$, we need translation pairs. We use the
same word-to-word bilingual lexicon mentioned in Section
\ref{crosslingual}. We then map the source vector space $S$ to the
bilingual space $\hat{S} = SW$ and use these embeddings.

\textbf{Baseline \barista.}  We also compare with the approach
proposed by \newcite{Gouws2015taskspecific}. The bilingual lexicon
used to create the pseudo-bilingual corpus is the same word-to-word
bilingual lexicon mentioned in Section \ref{crosslingual}. We follow
the authors' setup to create the pseudo-bilingual corpus. We create
bilingual embeddings by training skip-gram embeddings using the
Word2Vec toolkit on the pseudo-bilingual corpus using the same
parameters from Section \ref{embeddings}.

\textbf{Our method: BLSE.}
We implement our model \blse in Pytorch \cite{Pytorch} and initialize
the word embeddings with the pretrained word embeddings $S$ and $T$
mentioned in Section \ref{embeddings}. We use the word-to-word
bilingual lexicon from Section \ref{transdict}, tune the
hyperparameters $\alpha$, training epochs, and batch size on the
target development set and use the best hyperparameters achieved on
the development set for testing. ADAM \cite{Kingma2014a} is used in
order to minimize the average loss of the training batches.

\newpage
\textbf{Ensembles} We create an ensemble of \mt and each projection
method (\blse, \artetxe, \barista) by training a random forest
classifier on the predictions from \mt and each of these
approaches. This allows us to evaluate to what extent each projection
model adds complementary information to the machine translation
approach.

\begin{table}[b]
\newcommand{\sep}{\cmidrule(r){1-1}\cmidrule(rl){2-2}\cmidrule(lr){3-3}\cmidrule(lr){4-4}\cmidrule(rl){5-5}\cmidrule(rl){6-6}\cmidrule(rl){7-7}\cmidrule(l){8-8}}
\centering
\renewcommand*{\arraystretch}{0.8}
\setlength\tabcolsep{2.0mm}
\begin{tabular}{llrrrrrr}
\toprule
Model & & \multicolumn{1}{c}{\rt{voc}} & \multicolumn{1}{c}{\rt{mod}} & \multicolumn{1}{c}{\rt{neg}} & \multicolumn{1}{c}{\rt{know}}&\multicolumn{1}{c}{\rt{other}}&\multicolumn{1}{c}{\rt{\textit{total}}}\\
\sep\multirow{2}{*}{\footnotesize{\mt}}
 &bi & 49 & 26 & 19 & 14 & 5 & \textbf{113} \\
 &4 & 147 & 94 & 19 & 21 & 12 & \textbf{293} \\
\sep \multirow{2}{*}{\footnotesize{\artetxe}}
 &bi & 80 &44 &27 &14 &7 &\textbf{172}\\
 &4 & 182 &141 &19 &24 &19 &\textbf{385}\\
\sep \multirow{2}{*}{\footnotesize{\barista}}
 &bi & 89 &41 &27 &20 &7 &\textbf{184} \\ 
 &4 & 191 &109 &24 &31 &15 &\textbf{370} \\
\sep \multirow{2}{*}{\footnotesize{\blse}}
 &bi & 67 &45 &21 &15 &8 &\textbf{156}\\
 &4 & 146 &125 &29 &22 &19 &\textbf{341}\\
\bottomrule
\end{tabular}
\caption{Error analysis for different phenomena. See text for
  explanation of error classes.}
\label{erroranalysis}
\end{table}

\subsection{Results}
In Figure \ref{fig:results}, we report the results of all four
methods. Our method outperforms the other projection methods (the
baselines \artetxe and \barista) on four of the six experiments
substantially. It performs only slightly worse than the more
resource-costly upper bounds (\mt and \mono).
This is especially noticeable for the binary
classification task, where \blse performs nearly as well as machine
translation and significantly better than the other methods. We
perform approximate randomization tests \cite{Yeh2000} with 10,000
runs and highlight the results that are statistically significant (**p
$<$ 0.01, *p $<$ 0.05) in Table \ref{results}.

In more detail, we see that \mt generally performs better than the
projection methods (79--69 \F on binary, 52--44 on 4-class). \blse
(75--69 on binary, 41--30 on 4-class) has the best performance of the
projection methods and is comparable with \mt on the binary setup,
with no significant difference on binary Basque. \artetxe (67--46 on
binary, 35--21 on 4-class) and \barista (61--55 on binary, 40--34 on
4-class) are significantly worse than \blse on all experiments except
Catalan and Basque 4-class. On the binary experiment, \artetxe
outperforms \barista on Spanish (67.1 vs.\ 61.2) and Catalan (60.7
vs.\ 60.1) but suffers more than the other methods on the four-class
experiments, with a maximum \F of 34.9. \barista is relatively stable
across languages. 

\ensemble performs the best, which shows that \blse
adds complementary information to \mt. Finally, we note that all
systems perform successively worse on Catalan and Basque. This is
presumably due to the quality of the word embeddings, as well as the
increased morphological complexity of Basque.

\begin{figure}
\includegraphics[width = 3in]{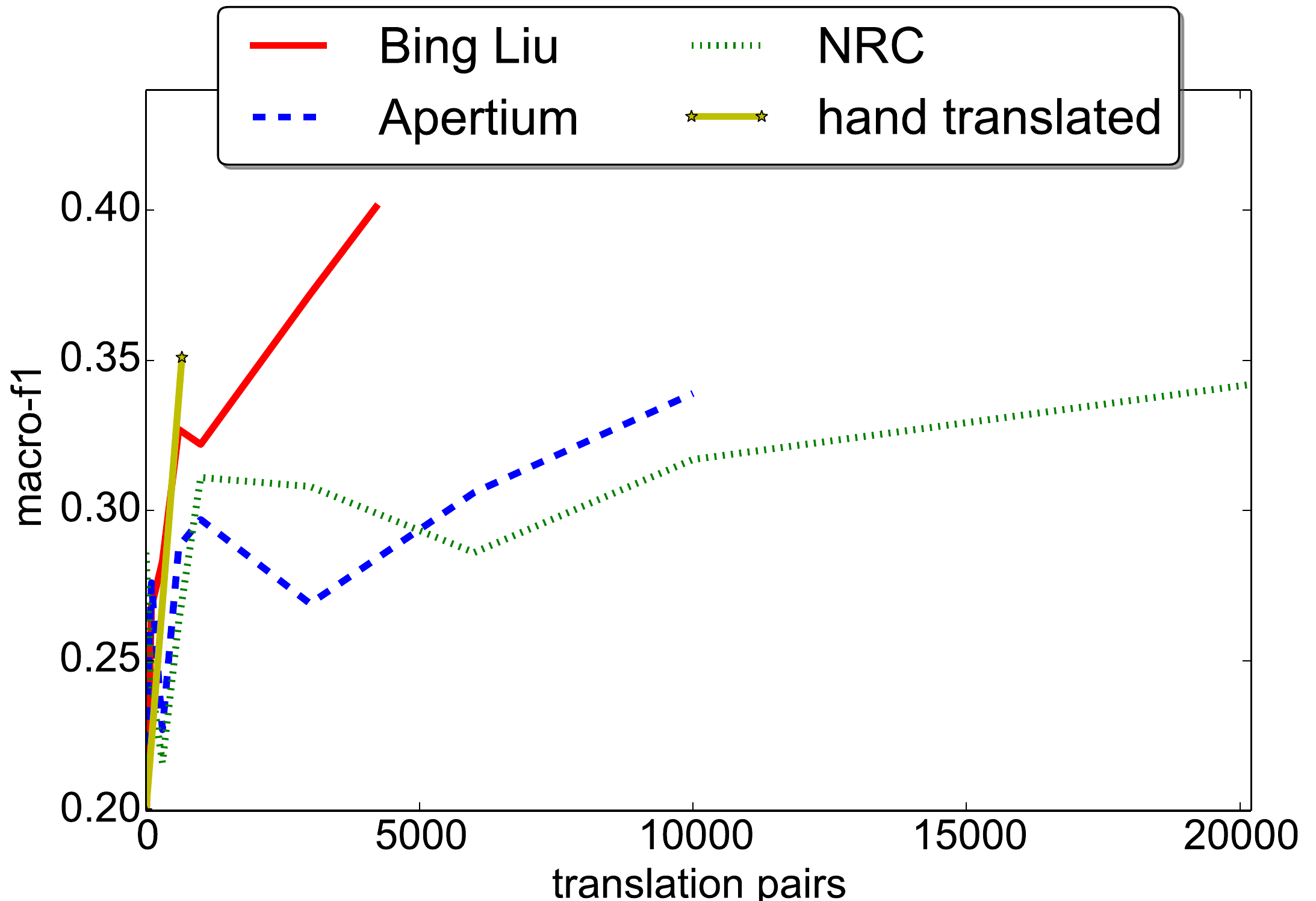}
\caption{Macro \F for translation pairs in the Spanish 4-class setup.
}
\label{fig:transdict}
\end{figure}

\begin{figure*}
\begin{center}
\includegraphics[width = 0.9\linewidth]{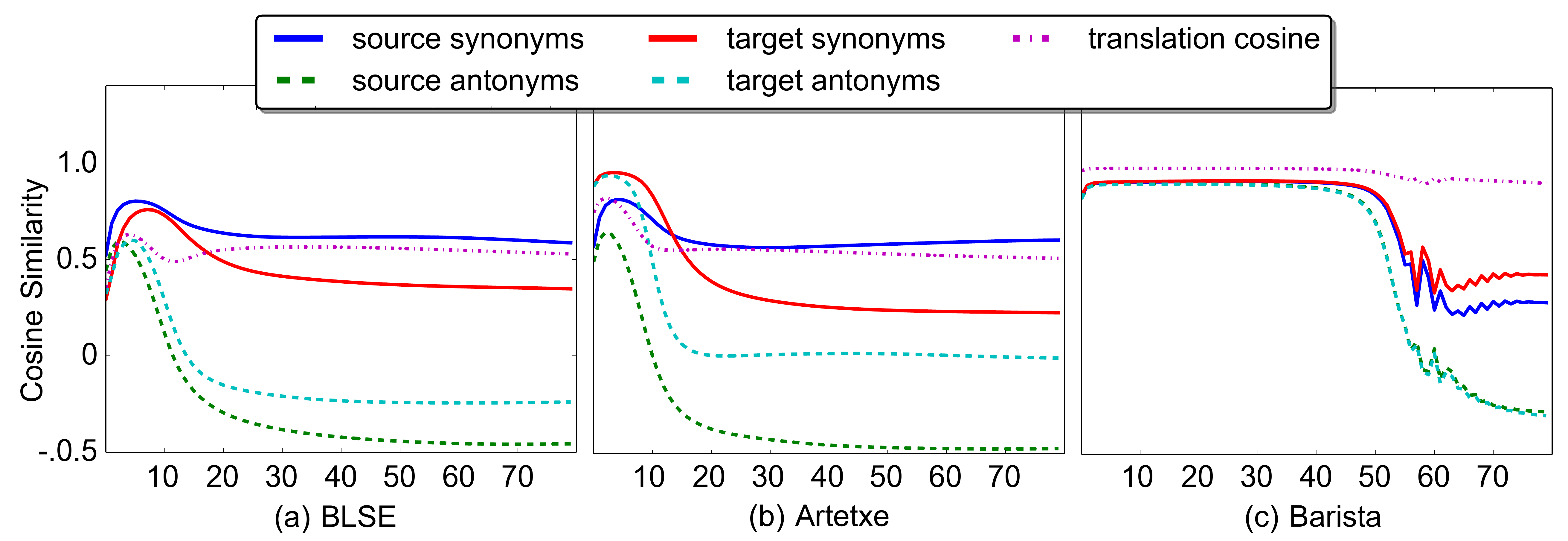}
\end{center}
\caption{Average cosine similarity between a subsample of translation
  pairs of same polarity (``sentiment synonyms'') and of opposing
  polarity (``sentiment antonyms'') in both target and source
  languages in each model. The x-axis shows
  training epochs. We see that \blse is able to learn that sentiment
  synonyms should be close to one another in vector space and
  sentiment antonyms should not.
}
\label{fig:synant}
\end{figure*}

\section{Model and Error Analysis}

We analyze three aspects of our model in further detail: (i) where most mistakes originate, (ii) the effect of the bilingual lexicon, and (iii) the effect and necessity of the target-language projection matrix $M'$. 

\subsection{Phenomena}

In order to analyze where each model struggles,
we categorize the mistakes and annotate all of the test
phrases with one of the following error classes: vocabulary (voc),
adverbial modifiers (mod), negation (neg), external knowledge (know) or other. Table \ref{erroranalysis}
shows the results.

\textbf{Vocabulary:} The most common way to express sentiment in hotel
reviews is through the use of polar adjectives (as in ``the room was
great) or the mention of certain nouns that are desirable (``it had a
pool''). Although this phenomenon has the largest total number of
mistakes (an average of 71 per model on binary and 167 on 4-class), it
is mainly due to its prevalence. \mt performed the best on the test examples which
according to the annotation require a correct understanding of the vocabulary
 (81 \F on binary /54 \F on 4-class),
with \blse (79/48) slightly worse. \artetxe (70/35) and \barista
(67/41) perform significantly worse. This suggests that \blse is better \artetxe and
\barista at transferring sentiment of the most important sentiment
bearing words.

\textbf{Negation:} Negation is a well-studied phenomenon in sentiment
analysis \cite{Pang2002,Wiegand2010,Zhu2014,Reitan2015}. Therefore, we
are interested in how these four models perform on phrases that
include the negation of a key element, for example ``In general, this
hotel isn't bad". We would like our models to recognize that the
combination of two negative elements ``isn't" and ``bad" lead to a
\textit{Positive} label.

Given the simple classification strategy,
all models perform relatively well on phrases with negation (all reach nearly
60 \F in the binary setting). However, while
\blse performs the best on negation in the binary setting (82.9
\F), it has more problems with negation in the 4-class setting
(36.9 \F).

\textbf{Adverbial Modifiers:} Phrases that are modified by an adverb,
\eg, the food was \textit{incredibly} good, are important for the
four-class setup, as they often differentiate between the base and
\textit{Strong} labels. In the binary case, all models reach more than
55 \F. In the 4-class setup, \blse only achieves 27.2 \F compared to
46.6 or 31.3 of \mt and \barista, respectively. Therefore, presumably,
our model does currently not capture the semantics of the target
adverbs well. This is likely due to the fact that it assigns too much
sentiment to functional words (see Figure \ref{fig:tsne}).

\textbf{External Knowledge Required:} These errors are difficult for
any of the models to get correct. Many of these include numbers which
imply positive or negative sentiment (350 meters from the beach is
\textit{Positive} while 3 kilometers from the beach is
\textit{Negative}). \blse performs the best (63.5 \F) while \mt
performs comparably well (62.5). \barista performs the worst (43.6).

\textbf{Binary vs. 4-class:} All of the models suffer when moving from
the binary to 4-class setting; an average of 26.8 in macro \F for \mt,
31.4 for \artetxe, 22.2 for \barista, and for 36.6 \blse. The two
vector projection methods (\artetxe and \blse) suffer the most,
suggesting that they are currently more apt for the binary setting.

\subsection{Effect of Bilingual Lexicon}

We analyze how the number of translation pairs affects our model. We train on the 4-class Spanish setup using the best hyper-parameters from the previous experiment. 

Research into projection techniques for bilingual word embeddings
\cite{Mikolov2013translation,Lazaridou2015,Artetxe2016} often uses a
lexicon of the most frequent 8--10 thousand words in English and their
translations as training data. We test this approach by taking the
10,000 word-to-word translations from the Apertium English-to-Spanish
dictionary\footnote{\url{http://www.meta-share.org}}. We also use the
Google Translate API to translate the NRC hashtag sentiment lexicon
\cite{Mohammad2013} and keep the 22,984 word-to-word
translations. We perform the same experiment as above and vary the
amount of training data from 0, 100, 300, 600, 1000, 3000, 6000,
10,000 up to 20,000 training pairs. Finally, we compile a small hand
translated dictionary of 200 pairs, which we then expand using target
language morphological information, finally giving us 657 translation
pairs\footnote{The translation took approximately one hour. We can
  extrapolate that hand translating a sentiment lexicon the size of
  the Bing Liu lexicon would take no more than 5 hours. }.  The macro
\F score for the Bing Liu dictionary climbs constantly with the
increasing translation pairs. Both the Apertium and NRC dictionaries
perform worse than the translated lexicon by Bing Liu, while the
expanded hand translated dictionary is competitive, as shown in
Figure~\ref{fig:transdict}.

While for some tasks, \eg, bilingual lexicon induction, using the most
frequent words as translation pairs is an effective approach, for
sentiment analysis, this does not seem to help. Using a translated
sentiment lexicon, even if it is small, gives better results.

\begin{figure}[t]
\centering
\includegraphics[width=0.9\linewidth]{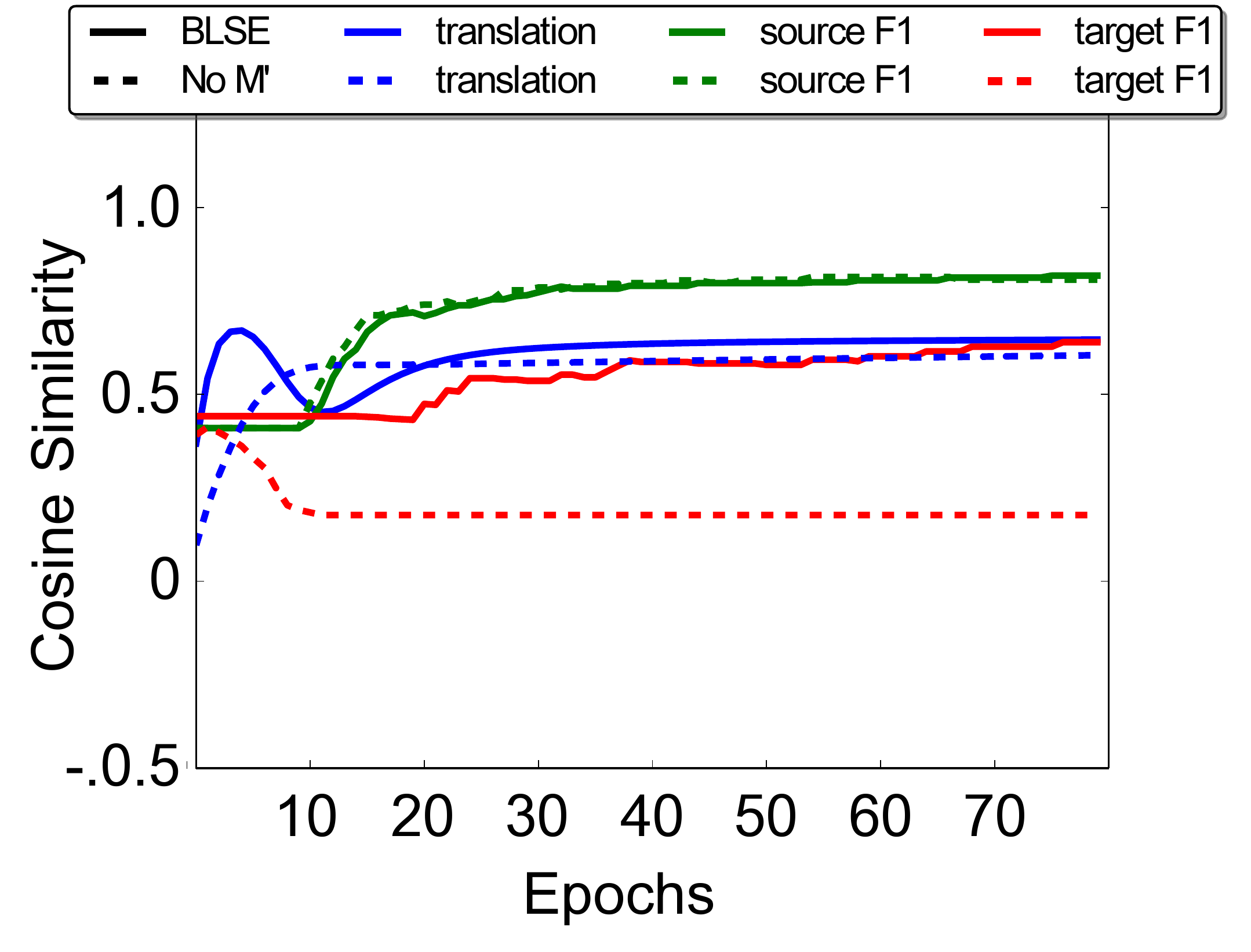}
\caption{\blse model (solid lines) compared to a variant without
  target language projection matrix $M'$ (dashed lines).
  ``Translation'' lines show the average cosine similarity between
  translation pairs. The remaining lines show \F scores for the source
  and target language with both varints of \blse. The modified model
  cannot learn to predict sentiment in the target language (red
  lines).  This illustrates the need for the second projection matrix
  $M'$.}
\label{fig:nomprime}
\end{figure}

\subsection{Analysis of $M'$}
\label{analysism}

The main motivation for using two projection matrices $M$ and $M'$ is
to allow the original embeddings to remain stable, while the projection
matrices have the flexibility to align translations and separate these
into distinct sentiment subspaces. To justify this design decision empirically, we perform
an experiment to evaluate the actual need for the target language
projection matrix $M'$: We create a simplified version of our model
without $M'$, using $M$ to project from the source to target and then
$P$ to classify sentiment.

The results of this model are shown in
Figure~\ref{fig:nomprime}. The modified model does learn to predict
in the source language, but not in the target language. This confirms
that $M'$ is necessary to transfer sentiment in our model.

\section{Qualitative Analyses of Joint Bilingual Sentiment Space}

In order to understand how well our model transfers sentiment
information to the target language, we perform two qualitative
analyses. First, we collect two sets of 100
positive sentiment words and one set of 100 negative sentiment
words. An effective cross-lingual sentiment classifier using embeddings should learn 
that two positive words should be closer in the shared bilingual 
space than a positive word
and a negative word. We test if \blse is able to do this by training our model and
after every epoch observing the mean cosine similarity between the
sentiment synonyms and sentiment antonyms after projecting to the
joint space.

We compare \blse with \artetxe and \barista by replacing the Linear
SVM classifiers with the same multi-layer classifier used in \blse and
observing the distances in the hidden layer. Figure~\ref{fig:synant}
shows this similarity in both source and target language, along with
the mean cosine similarity between a held-out set of translation pairs
and the macro \F scores on the development set for both source and
target languages for \blse, \barista, and \artetxe. From this plot, it
is clear that \blse is able to learn that sentiment synonyms should be
close to one another in vector space and antonyms should have a
negative cosine similarity. While the other models also learn this to
some degree, jointly optimizing both sentiment and projection gives
better results.

Secondly, we would like to know how well the projected vectors compare
to the original space. Our hypothesis is that some relatedness and
similarity information is lost during projection. Therefore, we
visualize six categories of words in t-SNE \cite{Vandermaaten2008}:
positive sentiment words, negative sentiment words, functional words,
verbs, animals, and
transport. 

\begin{figure}[t]
\centering
\includegraphics[width = 0.9\linewidth]{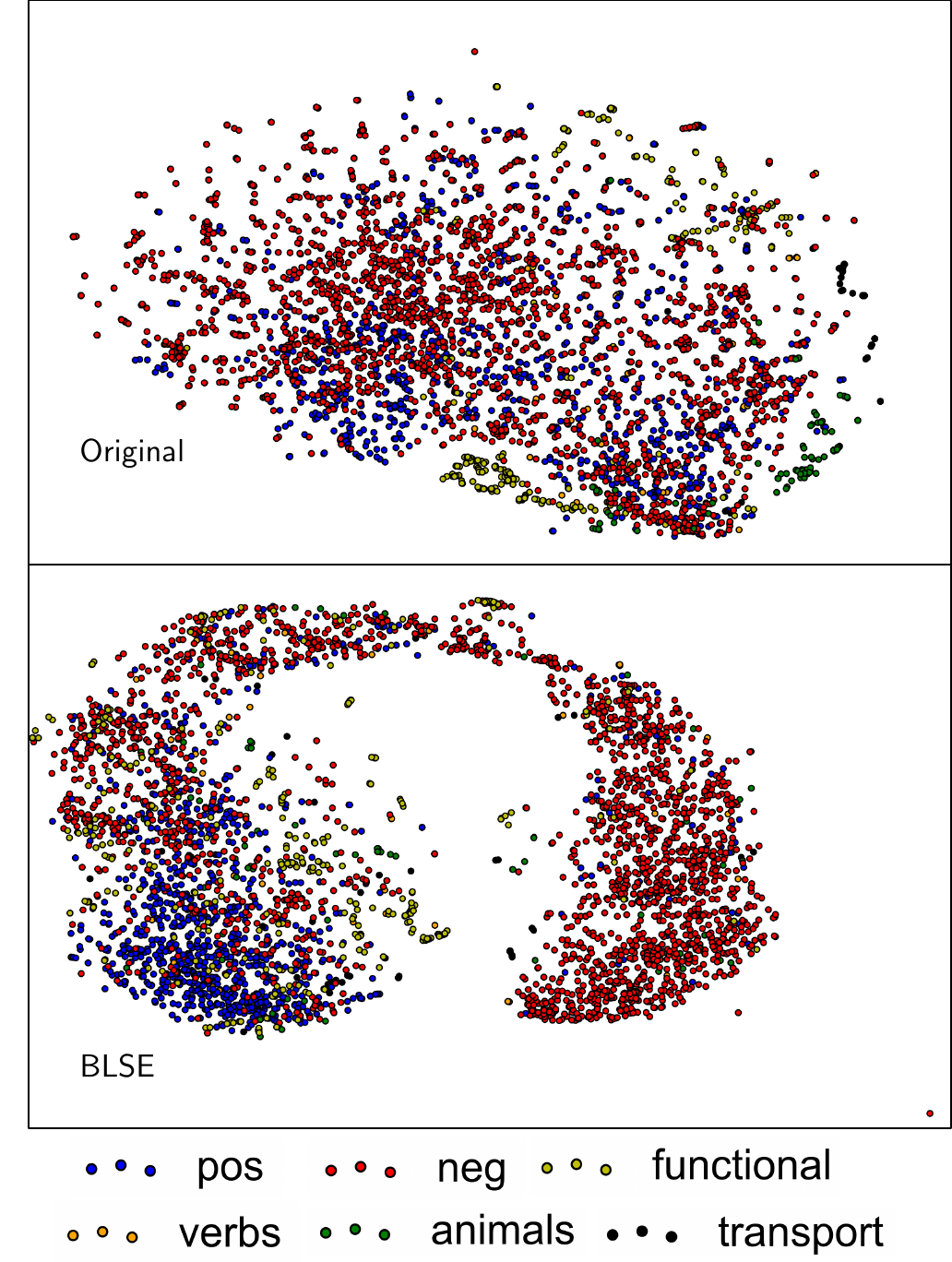}
\caption{t-SNE-based visualization of the Spanish vector space before
  and after projection with \blse. There is a clear separation of
  positive and negative words after projection, despite the fact that
  we have used no labeled data in Spanish.}
\label{fig:tsne}
\end{figure}

The t-SNE plots in Figure \ref{fig:tsne} show that the positive and
negative sentiment words are rather clearly separated after projection
in \blse. This indicates that we are able to incorporate sentiment
information into our target language without any labeled data in the
target language. However, the downside of this is that functional
words and transportation words are highly correlated with positive
sentiment.

\section{Conclusion}

We have presented a new model, \blse, which is able to leverage
sentiment information from a resource-rich language to perform
sentiment analysis on a resource-poor target language. This model
requires less parallel data than \mt and performs better than other
state-of-the-art methods with similar data requirements, an average of
14 percentage points in \F on binary and 4~pp on 4-class cross-lingual
sentiment analysis. We have also performed a phenomena-driven error
analysis which showed that \blse is better than \artetxe and \barista
at transferring sentiment, but assigns too much sentiment to
functional words. In the future, we will extend our model so that
it can project multi-word phrases, as well as single words, which
could help with negations and modifiers. 

\section*{Acknowledgements}
We thank Sebastian Padó, Sebastian Riedel, Eneko Agirre, and Mikel Artetxe for their conversations and feedback.

\bibliography{lit}

\begin{thebibliography}{}
\expandafter\ifx\csname natexlab\endcsname\relax\def\natexlab#1{#1}\fi

\bibitem[{Agerri et~al.(2014)Agerri, Bermudez, and Rigau}]{Agerri2014}
Rodrigo Agerri, Josu Bermudez, and German Rigau. 2014.
\newblock Ixa pipeline: Efficient and ready to use multilingual nlp tools.
\newblock In {\em Proceedings of the Ninth International Conference on Language
  Resources and Evaluation (LREC'14)\/}. pages 3823--3828.

\bibitem[{Agerri et~al.(2013)Agerri, Cuadros, Gaines, and Rigau}]{Agerri2013}
Rodrigo Agerri, Montse Cuadros, Sean Gaines, and German Rigau. 2013.
\newblock {OpeNER: Open polarity enhanced named entity recognition.}
\newblock {\em Sociedad Espa{\~{n}}ola para el Procesamiento del Lenguaje
  Natural\/} 51(Septiembre):215--218.

\bibitem[{Almeida et~al.(2015)Almeida, Pinto, Figueira, Mendes, and
  Martins}]{Almeida2015}
Mariana S.~C. Almeida, Claudia Pinto, Helena Figueira, Pedro Mendes, and
  Andr\'{e} F.~T. Martins. 2015.
\newblock Aligning opinions: Cross-lingual opinion mining with dependencies.
\newblock In {\em Proceedings of the 53rd Annual Meeting of the Association for
  Computational Linguistics and the 7th International Joint Conference on
  Natural Language Processing (Volume 1: Long Papers)\/}. pages 408--418.

\bibitem[{Artetxe et~al.(2016)Artetxe, Labaka, and Agirre}]{Artetxe2016}
Mikel Artetxe, Gorka Labaka, and Eneko Agirre. 2016.
\newblock Learning principled bilingual mappings of word embeddings while
  preserving monolingual invariance.
\newblock In {\em Proceedings of the 2016 Conference on Empirical Methods in
  Natural Language Processing\/}. pages 2289--2294.

\bibitem[{Artetxe et~al.(2017)Artetxe, Labaka, and Agirre}]{Artetxe2017}
Mikel Artetxe, Gorka Labaka, and Eneko Agirre. 2017.
\newblock Learning bilingual word embeddings with (almost) no bilingual data.
\newblock In {\em Proceedings of the 55th Annual Meeting of the Association for
  Computational Linguistics (Volume 1: Long Papers)\/}. pages 451--462.

\bibitem[{Balahur and Turchi(2014)}]{Balahur2014d}
Alexandra Balahur and Marco Turchi. 2014.
\newblock {Comparative experiments using supervised learning and machine
  translation for multilingual sentiment analysis}.
\newblock {\em Computer Speech {\&} Language\/} 28(1):56--75.

\bibitem[{Banea et~al.(2010)Banea, Mihalcea, and Wiebe}]{Banea2010}
Carmen Banea, Rada Mihalcea, and Janyce Wiebe. 2010.
\newblock Multilingual subjectivity: Are more languages better?
\newblock In {\em Proceedings of the 23rd International Conference on
  Computational Linguistics (Coling 2010)\/}. pages 28--36.

\bibitem[{Banea et~al.(2008)Banea, Mihalcea, Wiebe, and Hassan}]{Banea2008}
Carmen Banea, Rada Mihalcea, Janyce Wiebe, and Samer Hassan. 2008.
\newblock Multilingual subjectivity analysis using machine translation.
\newblock In {\em Proceedings of the 2008 Conference on Empirical Methods in
  Natural Language Processing\/}. pages 127--135.

\bibitem[{Barnes et~al.(2018)Barnes, Lambert, and Badia}]{Barnes2018}
Jeremy Barnes, Patrik Lambert, and Toni Badia. 2018.
\newblock Multibooked: A corpus of basque and catalan hotel reviews annotated
  for aspect-level sentiment classification.
\newblock In {\em Proceedings of 11th Language Resources and Evaluation
  Conference (LREC'18)\/}.

\bibitem[{Chandar et~al.(2014)Chandar, Lauly, Larochelle, Khapra, Ravindran,
  Raykar, and Saha}]{Chandar2014}
Sarath Chandar, Stanislas Lauly, Hugo Larochelle, Mitesh Khapra, Balaraman
  Ravindran, Vikas~C Raykar, and Amrita Saha. 2014.
\newblock An autoencoder approach to learning bilingual word representations.
\newblock In Z.~Ghahramani, M.~Welling, C.~Cortes, N.~D. Lawrence, and K.~Q.
  Weinberger, editors, {\em Advances in Neural Information Processing Systems
  27\/}, Curran Associates, Inc., pages 1853--1861.

\bibitem[{Chen et~al.(2016)Chen, Athiwaratkun, Sun, Weinberger, and
  Cardie}]{Chen2016}
Xilun Chen, Ben Athiwaratkun, Yu~Sun, Kilian~Q. Weinberger, and Claire Cardie.
  2016.
\newblock \href{http://arxiv.org/abs/1606.01614}{Adversarial deep averaging
  networks for cross-lingual sentiment classification}.
\newblock {\em CoRR\/} abs/1606.01614.
\newblock
  \href{http://arxiv.org/abs/1606.01614}{http://arxiv.org/abs/1606.01614}.

\bibitem[{Demirtas and Pechenizkiy(2013)}]{Demirtas2013}
Erkin Demirtas and Mykola Pechenizkiy. 2013.
\newblock {Cross-lingual polarity detection with machine translation}.
\newblock {\em Proceedings of the International Workshop on Issues of Sentiment
  Discovery and Opinion Mining - WISDOM '13\/} pages 9:1--9:8.

\bibitem[{Duh et~al.(2011)Duh, Fujino, and Nagata}]{Duh2011a}
Kevin Duh, Akinori Fujino, and Masaaki Nagata. 2011.
\newblock {Is machine translation ripe for cross-lingual sentiment
  classification?}
\newblock {\em Proceedings of the 49th Annual Meeting of the Association for
  Computational Linguistics: Human Language Technologies: short papers\/}
  2:429--433.

\bibitem[{Gouws et~al.(2015)Gouws, Bengio, and Corrado}]{Gouws2015}
Stephan Gouws, Yoshua Bengio, and Greg Corrado. 2015.
\newblock {BilBOWA: Fast bilingual distributed representations without word
  alignments}.
\newblock {\em Proceedings of The 32nd International Conference on Machine
  Learning\/} pages 748--756.

\bibitem[{Gouws and S{\o}gaard(2015)}]{Gouws2015taskspecific}
Stephan Gouws and Anders S{\o}gaard. 2015.
\newblock Simple task-specific bilingual word embeddings.
\newblock In {\em Proceedings of the 2015 Conference of the North American
  Chapter of the Association for Computational Linguistics: Human Language
  Technologies\/}. pages 1386--1390.

\bibitem[{Hermann and Blunsom(2014)}]{Hermann2014}
Karl~Moritz Hermann and Phil Blunsom. 2014.
\newblock Multilingual models for compositional distributed semantics.
\newblock In {\em Proceedings of the 52nd Annual Meeting of the Association for
  Computational Linguistics (Volume 1: Long Papers)\/}. Association for
  Computational Linguistics, Baltimore, Maryland, pages 58--68.

\bibitem[{Hu and Liu(2004)}]{Huandliu2004}
Minqing Hu and Bing Liu. 2004.
\newblock Mining opinion features in customer reviews.
\newblock In {\em Proceedings of the 10th ACM SIGKDD International Conference
  on Knowledge Discovery and Data Mining (KDD 2004)\/}. pages 168--177.

\bibitem[{Iyyer et~al.(2015)Iyyer, Manjunatha, Boyd-Graber, and
  Daume~III}]{Iyyer2015}
Mohit Iyyer, Varun Manjunatha, Jordan Boyd-Graber, and Hal Daume~III. 2015.
\newblock Deep unordered composition rivals syntactic methods for text
  classification.
\newblock In {\em Proceedings of the 53rd Annual Meeting of the Association for
  Computational Linguistics and the 7th International Joint Conference on
  Natural Language Processing (Volume 1: Long Papers)\/}. Beijing, China, pages
  1681--1691.

\bibitem[{Kingma and Ba(2014)}]{Kingma2014a}
Diederik Kingma and Jimmy Ba. 2014.
\newblock {Adam: A method for stochastic optimization}.
\newblock {\em Proceedings of the 3rd International Conference on Learning
  Representations (ICLR)\/} .

\bibitem[{Lazaridou et~al.(2015)Lazaridou, Dinu, and Baroni}]{Lazaridou2015}
Angeliki Lazaridou, Georgiana Dinu, and Marco Baroni. 2015.
\newblock {Hubness and pollution: delving into cross-space mapping for
  zero-shot learning}.
\newblock {\em Proceedings of the 53rd Annual Meeting of the Association for
  Computational Linguistics and the 7th International Joint Conference on
  Natural Language Processing\/} pages 270--280.

\bibitem[{Maas et~al.(2011)Maas, Daly, Pham, Huang, Ng, and Potts}]{Maas2011}
Andrew~L. Maas, Raymond~E. Daly, Peter~T. Pham, Dan Huang, Andrew~Y. Ng, and
  Christopher Potts. 2011.
\newblock Learning word vectors for sentiment analysis.
\newblock In {\em Proceedings of the 49th Annual Meeting of the Association for
  Computational Linguistics: Human Language Technologies\/}. pages 142--150.

\bibitem[{Meng et~al.(2012)Meng, Wei, Liu, Zhou, Xu, and Wang}]{Meng2012}
Xinfan Meng, Furu Wei, Xiaohua Liu, Ming Zhou, Ge~Xu, and Houfeng Wang. 2012.
\newblock \href{http://www.aclweb.org/anthology/P12-1060}{Cross-lingual mixture
  model for sentiment classification}.
\newblock In {\em Proceedings of the 50th Annual Meeting of the Association for
  Computational Linguistics (Volume 1: Long Papers)\/}. Association for
  Computational Linguistics, Jeju Island, Korea, pages 572--581.
\newblock
  \href{http://www.aclweb.org/anthology/P12-1060}{http://www.aclweb.org/anthology/P12-1060}.

\bibitem[{Mihalcea et~al.(2007)Mihalcea, Banea, and Wiebe}]{Mihalcea2007}
Rada Mihalcea, Carmen Banea, and Janyce Wiebe. 2007.
\newblock Learning multilingual subjective language via cross-lingual
  projections.
\newblock In {\em Proceedings of the 45th Annual Meeting of the Association of
  Computational Linguistics\/}. pages 976--983.

\bibitem[{Mikolov et~al.(2013)Mikolov, Le, and
  Sutskever}]{Mikolov2013translation}
Tomas Mikolov, Quoc~V. Le, and Ilya Sutskever. 2013.
\newblock Exploiting similarities among languages for machine translation.
\newblock {\em CoRR\/} abs/1309.4168.
\newblock {http://arxiv.org/abs/1309.4168}.

\bibitem[{Mohammad et~al.(2013)Mohammad, Kiritchenko, and Zhu}]{Mohammad2013}
Saif~M. Mohammad, Svetlana Kiritchenko, and Xiaodan Zhu. 2013.
\newblock Nrc-canada: Building the state-of-the-art in sentiment analysis of
  tweets.
\newblock In {\em Proceedings of the seventh international workshop on Semantic
  Evaluation Exercises (SemEval-2013)\/}.

\bibitem[{Pang et~al.(2002)Pang, Lee, and Vaithyanathan}]{Pang2002}
Bo~Pang, Lillian Lee, and Shivakumar Vaithyanathan. 2002.
\newblock Thumbs up? sentiment classification using machine learning
  techniques.
\newblock In {\em Proceedings of the ACL-02 Conference on Empirical methods in
  natural language processing-Volume 10\/}. Association for Computational
  Linguistics, pages 79--86.

\bibitem[{Paszke et~al.(2016)Paszke, Gross, Chintala, and Chanan}]{Pytorch}
Adam Paszke, Sam Gross, Soumith Chintala, and Gregory Chanan. 2016.
\newblock Pytorch deeplearning framework.
\newblock http://pytorch.org.
\newblock Accessed: 2017-08-10.

\bibitem[{Prettenhofer and Stein(2011)}]{Prettenhofer2011b}
Peter Prettenhofer and Benno Stein. 2011.
\newblock {Cross-lingual adaptation using structural correspondence learning}.
\newblock {\em ACM Transactions on Intelligent Systems and Technology\/}
  3(1):1--22.

\bibitem[{Rasooli et~al.(2017)Rasooli, Farra, Radeva, Yu, and
  McKeown}]{Rasooli2017}
Mohammad~Sadegh Rasooli, Noura Farra, Axinia Radeva, Tao Yu, and Kathleen
  McKeown. 2017.
\newblock Cross-lingual sentiment transfer with limited resources.
\newblock {\em Machine Translation\/} .

\bibitem[{Reitan et~al.(2015)Reitan, Faret, Gamb\"{a}ck, and
  Bungum}]{Reitan2015}
Johan Reitan, J{\o}rgen Faret, Bj\"{o}rn Gamb\"{a}ck, and Lars Bungum. 2015.
\newblock Negation scope detection for twitter sentiment analysis.
\newblock In {\em Proceedings of the 6th Workshop on Computational Approaches
  to Subjectivity, Sentiment and Social Media Analysis\/}. pages 99--108.

\bibitem[{Tang et~al.(2014)Tang, Wei, Yang, Zhou, Liu, and Qin}]{Tang2014}
Duyu Tang, Furu Wei, Nan Yang, Ming Zhou, Ting Liu, and Bing Qin. 2014.
\newblock Learning sentiment-specific word embedding for twitter sentiment
  classification.
\newblock In {\em Proceedings of the 52nd Annual Meeting of the Association for
  Computational Linguistics (Volume 1: Long Papers)\/}. pages 1555--1565.

\bibitem[{Van~der Maaten and Hinton(2008)}]{Vandermaaten2008}
Laurens Van~der Maaten and Geoffrey Hinton. 2008.
\newblock {Visualizing data using t-sne}.
\newblock {\em Journal of Machine Learning Research\/} 9:2579--2605.

\bibitem[{Wan(2009)}]{Wan2009}
Xiaojun Wan. 2009.
\newblock Co-training for cross-lingual sentiment classification.
\newblock In {\em Proceedings of the Joint Conference of the 47th Annual
  Meeting of the ACL and the 4th International Joint Conference on Natural
  Language Processing of the AFNLP\/}. pages 235--243.

\bibitem[{Wiegand et~al.(2010)Wiegand, Balahur, Roth, Klakow, and
  Montoyo}]{Wiegand2010}
Michael Wiegand, Alexandra Balahur, Benjamin Roth, Dietrich Klakow, and
  Andr\'es Montoyo. 2010.
\newblock A survey on the role of negation in sentiment analysis.
\newblock In {\em Proceedings of the Workshop on Negation and Speculation in
  Natural Language Processing\/}. pages 60--68.

\bibitem[{Xiao and Guo(2012)}]{Xiao2012}
Min Xiao and Yuhong Guo. 2012.
\newblock Multi-view adaboost for multilingual subjectivity analysis.
\newblock In {\em Proceedings of COLING 2012\/}. pages 2851--2866.

\bibitem[{Yeh(2000)}]{Yeh2000}
Alexander Yeh. 2000.
\newblock More accurate tests for the statistical significance of result
  differences.
\newblock In {\em Proceedings of the 18th Conference on Computational
  linguistics (COLING)\/}. pages 947--953.

\bibitem[{Zhou et~al.(2016)Zhou, Zhu, He, and Hu}]{Zhou2016}
Guangyou Zhou, Zhiyuan Zhu, Tingting He, and Xiaohua~Tony Hu. 2016.
\newblock Cross-lingual sentiment classification with stacked autoencoders.
\newblock {\em Knowledge and Information Systems\/} 47(1):27--44.

\bibitem[{Zhou et~al.(2015)Zhou, Chen, Shi, and Huang}]{Zhou2015}
HuiWei Zhou, Long Chen, Fulin Shi, and Degen Huang. 2015.
\newblock Learning bilingual sentiment word embeddings for cross-language
  sentiment classification.
\newblock In {\em Proceedings of the 53rd Annual Meeting of the Association for
  Computational Linguistics and the 7th International Joint Conference on
  Natural Language Processing (Volume 1: Long Papers)\/}. pages 430--440.

\bibitem[{Zhu et~al.(2014)Zhu, Guo, Mohammad, and Kiritchenko}]{Zhu2014}
Xiaodan Zhu, Hongyu Guo, Saif Mohammad, and Svetlana Kiritchenko. 2014.
\newblock An empirical study on the effect of negation words on sentiment.
\newblock In {\em Proceedings of the 52nd Annual Meeting of the Association for
  Computational Linguistics (Volume 1: Long Papers)\/}. pages 304--313.

\end{thebibliography}
\bibliographystyle{acl_natbib}

\end{document}